\documentclass[letterpaper, 10 pt, conference]{ieeeconf}  

\IEEEoverridecommandlockouts                         
\pdfoutput=1

\title{\LARGE\bf Contact Surface Estimation via Haptic Perception}
  
\author{
  Hsiu-Chin Lin$^{1}$ and Michael Mistry$^2$
  \thanks{Hsiu-Chin Lin$^1$ ({\tt\small hsiu-chin.lin@cs.mcgill.ca}) is at the School of Computer Science, McGill University, Canada and the Department of Electrical and Computer Engineering, McGill University, Canada. Michael Mistry$^2$ is at the School of Informatics, University of Edinburgh, UK.}  
}

\usepackage{times}
\usepackage{graphicx}\graphicspath{{fig/}} 
\usepackage{epstopdf}
\usepackage{algorithm}
\usepackage{algorithmic}
\usepackage{multirow} 
\usepackage{multicol}
\usepackage{array}
\usepackage{mdwmath}
\usepackage{mdwtab}
\usepackage{rotating}
\usepackage{caption}
\usepackage{subfig}
\usepackage{amssymb}
\usepackage{verbatim}
\usepackage[cmex10]{amsmath}
\let\labelindent\relax 
\usepackage{mathtools}

\usepackage[utf8]{inputenc}
\usepackage[english]{babel}

\mathchardef\mhyphen="2D   

\newcommand{\R}     {\mathbb{R}}          
\newcommand{\T}     {\top}                

\newcommand{\vectornorm}[1]{||#1||}

\newcommand{\bx}     {\mathbf{x}}         

\newcommand{\bA}     {\mathbf{A}}         
\newcommand{\bc}     {\mathbf{c}}         
\newcommand{\bv}     {\mathbf{v}}         
\newcommand{\bq}     {\mathbf{q}}         
\newcommand{\bJ}     {\mathbf{J}}         
\newcommand{\bqdot}  {\dot{\bq}}          

\newcommand{\bqddot} {\ddot{\bq}}         
\newcommand{\btau}   {\boldsymbol{\tau}}  
\newcommand{\bK}     {\mathbf{K}}         
\newcommand{\bB}     {\mathbf{B}}         

\renewcommand{\nu}  {q}                   

\newcommand{\bX}     {\mathbf{X}}         

\renewcommand{\nu}      {q}                      
\newcommand{\bD}      {\mathbf{D}}             %

\newcommand{\bLambda}     {\boldsymbol{\Lambda}}       %

\DeclareMathOperator{\Jacobian}{{\mathbf J} }

\DeclareMathOperator{\Identity}{{\mathbf I} }

\DeclareMathOperator{\bh }{\mathbf{h}}  
 
\DeclareMathOperator{\bZero }{\mathbf{0}} 
\DeclareMathOperator{\bU }{\mathbf{U}} 
\DeclareMathOperator{\bV }{\mathbf{V}}

\newcommand{\bM}{\mathbf{M}}

\newcommand{\prob}[1]{\mathcal{P}(#1)}

\newcommand{\contactPoints}{\mathbf{c}}
\newcommand{\contactVectors}{\mathbf{v}}
\newcommand{\contactVectorSet}{\mathbf{D}}
\newcommand{\dimPoints} {\mathcal{K}}
\newcommand{\dimVectors} {\mathcal{V}}

\newcommand{\bp} {\mathbf{p}}

\newcommand{\bS} {\mathbf{S}}
\newcommand{\contactNormal}{\boldsymbol{\hat{n}}}

\newcommand{\bpdot}{\dot{\bp}}
\newcommand{\bpddot}{\ddot{\bp}}
\newcommand{\bP}{\mathbf{P}}
\newcommand{\bJdot}{\dot{\mathbf{J}}}

\newcommand{\dimFeet}{3}
\newcommand{\exploreMag}{\alpha}
\newcommand{\exploreAng}{\theta}

\newcommand{\obsAve}{\mu_D}
\newcommand{\obsVar}{\sigma^2_D}

\newcommand{\NormalDist}[2]{\mathcal{N}(#1,#2)}
\newcommand{\stoppingThreshold}{\epsilon_\Lambda}
\newcommand{\contactForce}{\boldsymbol{\lambda}_c}
\newcommand{\contactForcePerLeg}{\boldsymbol{\lambda}_{c,i}}

\newcommand{\expForce}{\mathbf{f}_w}
\newcommand{\expTorque}{\btau_w}
\newcommand{\expJacobian}{\Jacobian_w}
\newcommand{\expNormal}{\mathbf{f}_\perp}
\newcommand{\expTangent}{\mathbf{f}_\parallel}

\newcommand{\contactSurface}{\mathcal{S}}
\newcommand{\contactFriction}{\mu}

\newcommand{\obsFriction}{z}

\newcommand{\NormalGammaDist}[1]{\sim\mathcal{N}\mathcal{G}(#1)}
\newcommand{\obsPre}{\varrho}

\newcommand{\NormalGammaAve}{\tilde{\boldsymbol{n}}}       	
\newcommand*{\sref}[1]{~\S\ref{s:#1}}            
\newcommand*{\fref}[1]{\figurename~\ref{f:#1}}  
\newcommand*{\eref}[1]{(\ref{e:#1})}            

\usepackage[inline]{enumitem}
\setlist{nolistsep}

\newenvironment{example}[1][Example]{\begin{trivlist}
\item[\hskip \labelsep {\bfseries #1}]}{\end{trivlist}}
\newenvironment{remark}[1][Remark]{\begin{trivlist}
\item[\hskip \labelsep {\bfseries #1}]}{\end{trivlist}}

\usepackage{color}
\newcommand{\todo}[1]{\textcolor{magenta}{(Todo:~#1)}}

\usepackage[percent]{overpic} 
\makeatletter\newcommand{\manuallabel}[2]{\def\@currentlabel{#2}\label{#1}}\makeatother

\begin{document}

\maketitle
\thispagestyle{empty}
\pagestyle{empty}

								\begin{abstract}
Legged systems need to optimize contact force in order to maintain contacts.
For this, the controller needs to have the knowledge of the surface geometry and how slippery the terrain is.
We can use a vision system to realize the terrain, but the accuracy of the vision system degrades in harsh weather, and it cannot visualize the terrain if it is covered with water or grass. 
Also, the degree of friction cannot be directly visualized.
In this paper, we propose an online method to estimate the surface information via haptic exploration.
We also introduce a probabilistic criterion to measure the quality of the estimation.
The method is validated on both simulation and a real robot platform.
\end{abstract}

\section{Introduction}			\label{s:introduction} 	
\noindent Legged robots have many practical applications in both industry and the public sector; particularly, in places that are too dangerous for humans, such as mining inspections in underground tunnels or the aftermath of an earthquake.
The need for legged systems instead of wheeled systems is apparent since the passages are normally uneven and difficult to maneuver.

Yet, vision systems may degrade in harsh conditions, such as in dusty and foggy environments. 
Moreover, vision systems are unable to detect/realize terrain under cover (e.g., A LiDAR system may not be able to distinguish a water surface or high grass from a solid contact.)
The challenge here is how can the robot maneuver over an unknown environment when the perception system fails? 

Model-based control has shown promising results in the robotic community, 
ranging from bi-manual manipulation~\cite{Lin.2018}, quadruped locomotion~\cite{Kim.2018}, to whole-body humanoid control~\cite{Dietrich.2012}\cite{Tedrake.2016.AR}.
This control mechanism is particularly important for legged locomotion, where the robots need to maintain appropriate contact forces at the point-of-contacts to avoid slippage. However, those methods require the {\em terrain friction} and {\em surface geometry} to be provided a priori, since this information is needed for contact force optimization~\cite{Nori.2015}\cite{Bellicoso.2018}. 
Additionally, the surface properties are also required for footstep planning over rough terrain~\cite{Schaal.2009}\cite{Tedrake.2016.Humanoids}.

The surface geometry can be gathered via vision system~\cite{Fankhauser.2018}\cite{Belter.2019}; on the other hand, the accuracy can be disturbed in harsh weather conditions. 
Besides, the surface friction can not be directly visualized.
Without sensing skills or prior information about the environment, this framework could only be applied successfully in laboratory settings.

One potential solution is to utilize a single foot as a {\em haptic sensor} while standing on the other three. 
This concept was motivated by how humans explore the environment. 
For example, when entering a dark room for the first time, we use our hand to touch our surroundings to avoid bumping into a wall, before making a move.
The robot can use a single foot to touch the surface near the next desired footstep position and estimate the terrain information. 

Previous work on estimation of surface normal takes the {\em learning by demonstration} approach, where the users move the robot on the contact surface to sense the constraint, and then estimate the constraint {\em offline}~\cite{Lin.2017}\cite{Subramani.2018}\cite{Armesto.2018}\cite{Amanhoud.2019}. 
However, it is not practical to provide demonstrations to all environments for the robots where the constraints vary. Instead of learning from human demonstration, in~\cite{Ortenzi.2016}, the robot performs a set of trajectories to collect the motion data. Nevertheless, the data collection took 2 minutes in this work, which is also not practical in real robotic applications.

We can also find work related to the online estimation of terrain information in the literature of quadruped locomotion.~\cite{Homberger.2019} estimates the height of the support surface using proprioception and LiDAR, and~\cite{Gehring.2015} proposed a method to find the surface normal under the assumption that terrain is consistent for all standing legs. In contrast, our work relaxes such assumptions.

Additionally, the estimation of the friction coefficient has been carried out in many works. While most measures the contact force using force-torque sensor~\cite{Cutkosky.1993}\cite{Nakamura.2001}, no sensor is required in our work. On top of that, previous work only attempted to find the friction coefficient on a flat surface.

\begin{figure}
  \centering
  \includegraphics[width=0.25\textwidth, angle=180]{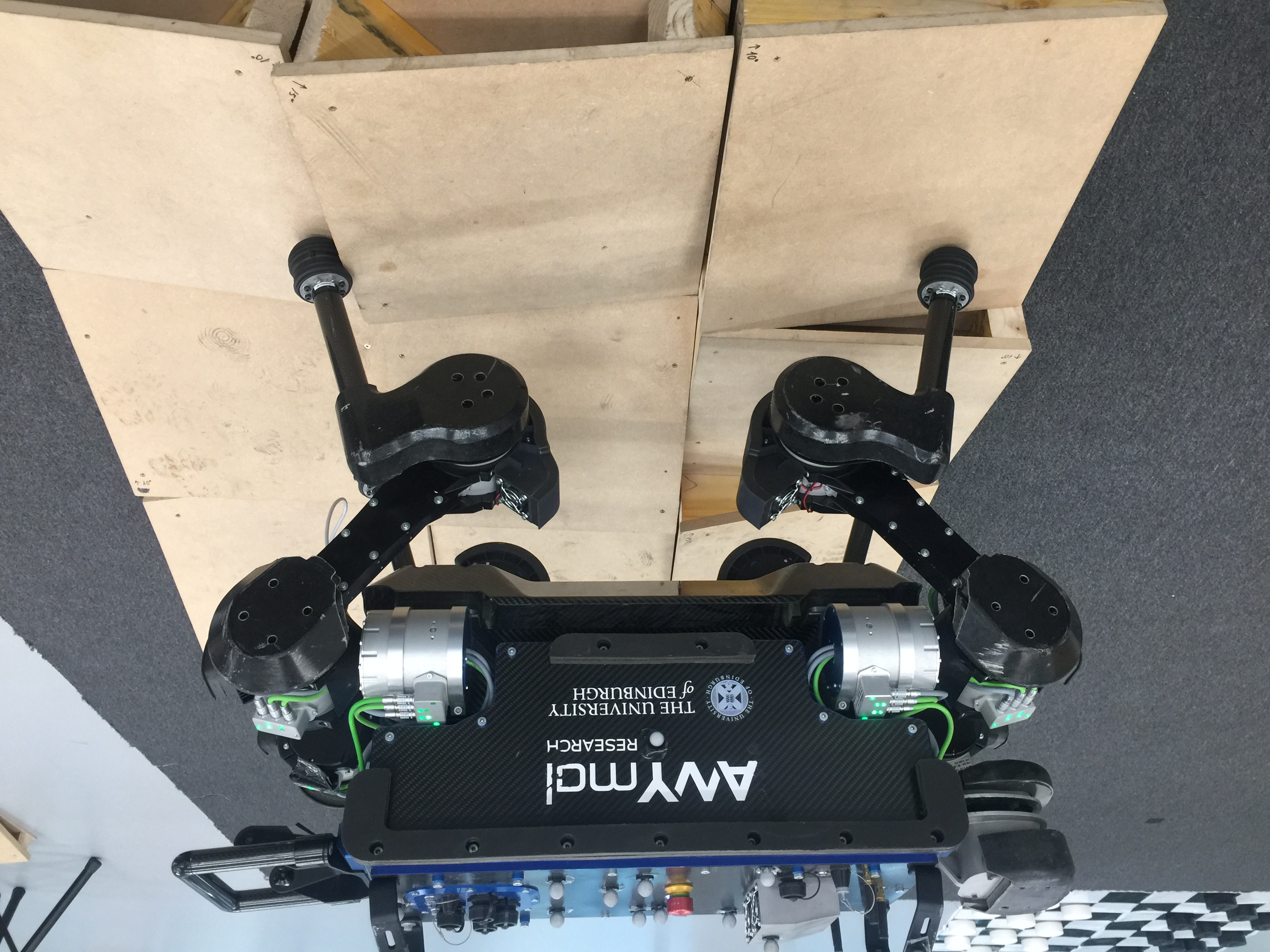}
  \caption{ANYmal robot walking over an uneven terrain}
  \label{f:intro}
\end{figure}
In this paper, we propose a method to estimate the surface properties when the robot loses information about the terrain. 
The contribution of this paper includes 
(1) an efficient online method that estimates the surface normal,
(2) a haptic sensing strategy that senses the environment by the robot itself, and 
(3) a probabilistic criterion that evaluates the result of the estimation, so the robot can stop exploring when it is confident about the estimation.
These methods have been evaluated in simulation and on ANYmal~\cite{Hutter.2016} (see~\fref{intro}).

\section{Background} 			\label{s:background} 	\noindent
First, we define the problem and clarify what quantity we aim to estimate in~\sref{background-problem} and briefly review the Bayesian inference method in~\sref{backgroun-bayes}. 
Finally, we highlight our contribution in~\sref{background-contribution}.

\subsection{Problem Definition}
\label{s:background-problem}
\noindent 
Let $\bq, \bqdot, \bqddot \in \R^{d+6}$ be the generalized positions, velocities, and accelerations of a floating-based $n$ degree-of-freedom robotics system, the rigid body dynamics with $k$ contacts can be described as 
\begin{equation}
 \bM \bqddot + \bh = \bB\btau + \sum_{i}^k\Jacobian_{c,i}^\T \contactForcePerLeg
 \label{e:background-dynamics}
\end{equation}
\noindent 
where 
$\bM\in\R^{(d+6)\times(d+6)}$ is the inertia matrix,
$\bh\in\R^{(d+6)}$ is the Coriolis, centrifugal, and gravitational force,
$\bB \in\R^{(d+6)\times d}$ is the selection matrix, 
$\btau \in\R^d$ is the control torque,
$\Jacobian_{c,i} \in\R^{3\times (d+6)}$ is the constraint Jacobian that relates the joint space to the $i^{th}$ stance foot, 
and $\contactForcePerLeg\in\R^{3}$ is the reaction force at the $i^{th}$ contact.

For model-based controllers, such as~\cite{Kim.2018}\cite{Bellicoso.2018}, the goal is to find the control torques for the desired tasks while satisfying a set of kinematics and dynamics constraints.
A typical constraint that avoids robots from slippage is the friction cone constraints~\cite{Trinkle.1997}. 
Given a unit vector $\contactNormal$ normal to the contact surface, we can divide the contact force $\contactForce$ into three components such that 
$\lambda_n=\contactNormal^\T\contactForce$ is normal to the contact surface, and 
$\lambda_x, \lambda_y$ are tangential to the contact surface. 
By Coulomb's Law, the magnitude of the tangential force should not exceed the friction coefficient $\mu$ times the magnitude of the normal force to avoid slipping
  \begin{equation}  
    \mu \lambda_{z} \geq \sqrt{\lambda_{x}^2 + \lambda_{y}^2 } 
    \label{e:background-friction}
  \end{equation}

\noindent where $\mu$ is the friction coefficient which depends on the material of the surface. 

An example can be seen in~\fref{background-friction-cone}. 
The surface normal $\contactNormal$ determines the direction of the friction cone, and the friction coefficient $\contactFriction$ determines the shape of the friction cone.
Geometrically, the controller needs to find the torque $\btau$ such that the resulting contact force $\contactForce$ lies with the friction cone.

\begin{figure}
  \centering
  \includegraphics[width=0.25\textwidth]{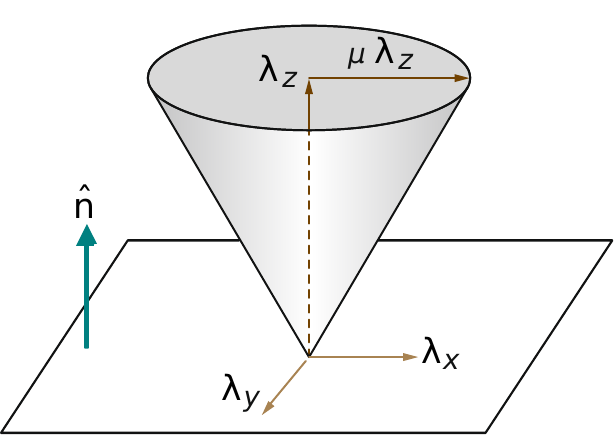}
  \caption{An illustration of the friction cone constraint. The surface normal $\contactNormal$ controls the direction of the cone, and the friction coefficient $\mu$ controls the size of the cone.}
  \label{f:background-friction-cone}
\end{figure}
Ideally, we need to provide the $\contactNormal$ and $\contactFriction$ to the controller before the optimization of control torques. 
Instead of providing this information to the controller a priori, the goal of this paper is to explore and estimate these values by the robot itself.

\subsection{Bayesian inference}
\label{s:backgroun-bayes}
\noindent Given a random variable $\bX$ with unknown distribution and a set of observations $\bD$, Bayesian inference~\cite{Murphy.2012} is a method to infer the probability of the unknown variables $\bX$.
The {\em prior distribution} $\prob{\bX}$ is the distribution of the parameters before the observations.
The {\em likelihood} $\prob{\bD|\bX}$ is the distribution of the observations, conditional on the variable.
The {\em posterior probability} $\prob{\bX|\bD}$ is the conditional probability after some observations, which is determined by Bayes' rule
\[
\prob{\bX|\bD} \propto \prob{\bD|\bX} \prob{\bX} 
\]

\noindent If the prior and posterior have the same distribution, the prior is the {\em conjugate prior} of the likelihood distribution. 

In this work, we would like to know the probability of our estimation $\contactNormal$ and $\contactFriction$ without knowing the true distribution of these two variables. In~\sref{normal-probability} and~\sref{friction-probability}, we will use Bayesian inference with conjugate prior to approximate the probability of our estimation.

\subsection{Contributions} \label{s:background-contribution}
\noindent 
In this paper, our goal is to estimate $\contactNormal$ and $\contactFriction$ with minimum prior information.
The contributions of this paper include the following:




\begin{itemize}
 \item Previous work on estimation of $\contactNormal$ takes the offline learning approach~\cite{Lin.2017}\cite{Subramani.2018}\cite{Armesto.2018}, while we introduced an efficient {\em online}.
 
 \item In~\cite{Ortenzi.2016}, the robot explores the environment by sliding on the contact surface. This work took 2 minutes to finish one estimation since it was hard to slide on an unknown surface. Here, the robot explores by touching and finishes in a realistic time frame.
 
 \item Instead of using force/torque sensor to estimate $\contactFriction$~\cite{Cutkosky.1993}\cite{Nakamura.2001}, our method does not require the F/T sensor, which tends to be noisy.
 
 \item While previous work cannot quantify how well the estimation is,
    we proposed a probabilistic criterion that evaluates the results of the estimation, so the robot may stop exploring when it is confident about the estimation.
    
\end{itemize}

\section{Surface Normal Estimation}	\label{s:method}	\noindent In this section, we describe our method for surface normal estimation.
When the robot loses its visual accuracy, we will collect data of the surface via {\em haptic exploration}.
Namely, we use a single foot as a haptic sensor while standing on the other three.

\subsection{Data Collection via Haptic Exploration}

\noindent 
Let $\bp, \bpdot, \bpddot \in \R^{\dimFeet}$ denotes the position, velocities, and accelerations of the feet.
The joint space and end-effector space are related through the Jacobian 
$\bpdot = \bJ(\bq)\bqdot$ and $\bpddot = \bJ(\bq)\bqddot + \bJdot \bqdot$.

Let $\bp^* \in \R^{\dimFeet}$ be the desired foot position for the current swing leg according to some predefined footstep planner. 
Before making that contact and shift the center-of-mass, the swing foot tries to {\em explore} the area around $\bp^*$  to sense the area.
An example can be found in~\fref{exploration-next-footstep}, the left front leg is the swing leg, and the red point is the desired foot position $\bp^*$ for this leg. 

\begin{figure}
  \centering
  \includegraphics[width=0.35\textwidth]{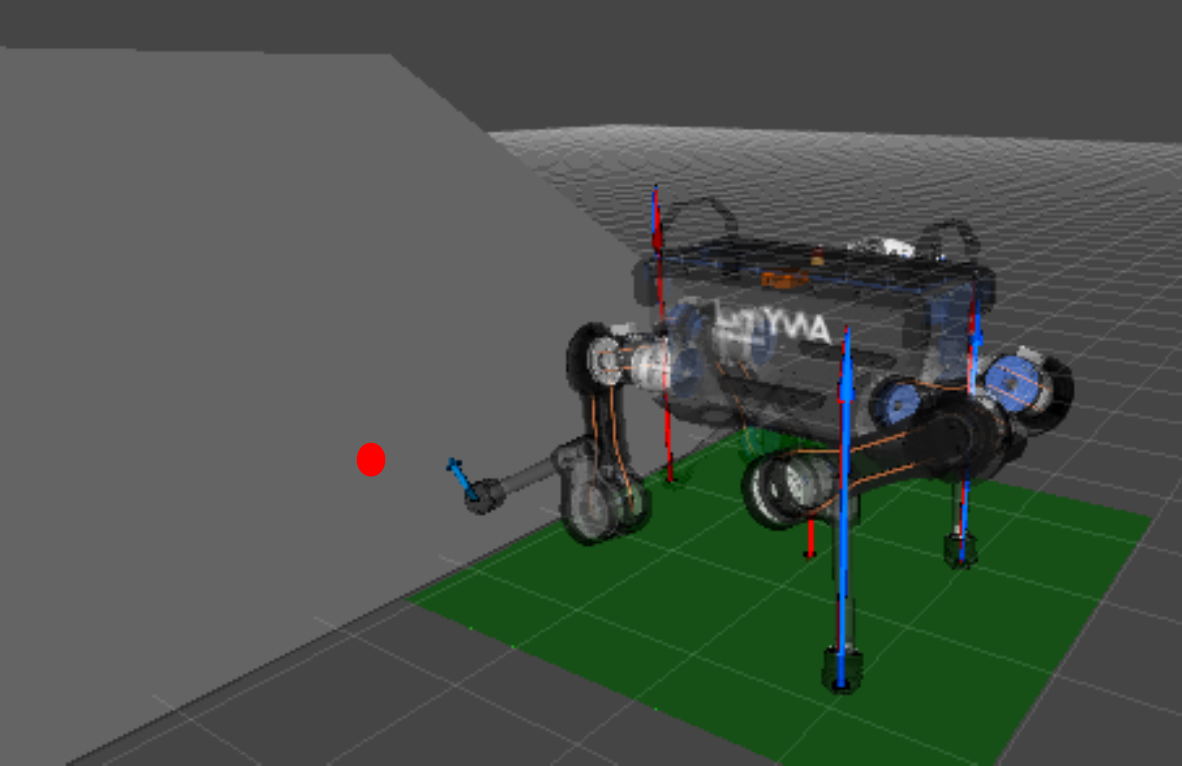}
  \caption{ 
  The left front leg is the swing leg, and the red point is the desired foot position $\bp^*$  for this leg.
  } 
  \label{f:exploration-next-footstep}
\end{figure}

The robot explores the area around $\bp^*$ by making $\dimPoints$ contacts with that area according to
\begin{equation}
\begin{aligned}
 \bp^*_k &= \bp^* + \delta_k  \\
\delta_k &=  \left[ \exploreMag\cos(\exploreAng_k), \exploreMag\sin(\exploreAng_k), z \right]^\T
\end{aligned}
\label{e:surface-exploration-points}
\end{equation}
where 
$\exploreMag$ defines the magnitude of $\delta_i$ (i.e., how far it is from the $\bp^*$) and 
$\exploreAng_k$ defines the angle between $[1,0]$ and $\delta_k$.
In an ideal world, only $\dimPoints=3$ points are needed to define the normal vector. In reality, more contact points will be needed due to sensory noise.

\begin{example} 
Let $\bp^*=\left[0,0,0\right]^\T$ be the desired position for the n/ext footstep. We use~\eref{surface-exploration-points} to set the exploration points. 
Let $\exploreMag=1$ and ${\exploreAng_k = \left[0, 90, 180, 270\right]}$, the robot will try to make contact at 
$\bp^*_1 =\left[ 1,0,0 \right]^\T$,
$\bp^*_2 =\left[-1,0,0 \right]^\T$,
$\bp^*_3 =\left[ 0,-1,0 \right]^\T$, and
$\bp^*_4 =\left[ 0,1,0 \right]^\T$.
\end{example}

\noindent 
For each $k$ exploration, when the robot detects collision between the swing foot and the contact surface, the contact positions of the foot $\bc \in \R^3$ are recorded.
(Note that, $\bc$ may not be identical with $\bp$, since there might be an early collision and/or tracking error)
A set of $\dimPoints$ data points will be collected $\bP = \left[\bc_1, \bc_2, \dots, \bc_k \right]$.
An example of exploration can be found in~\fref{exploration-contact-points}.

In the above example, the magnitude of the exploration is set to $\exploreMag=1$. In practice, a reasonable choice of $\exploreMag$ is proportional to the size of the foot. 
Recall that, the reason for gathering the terrain properties is to optimize the contact force for the stance legs, and the only information that matters is the terrain underneath each foot. This simplifies the complexity of dealing with highly uneven terrain.

\begin{figure}[t!]
\centering
\includegraphics[width=.65\linewidth]{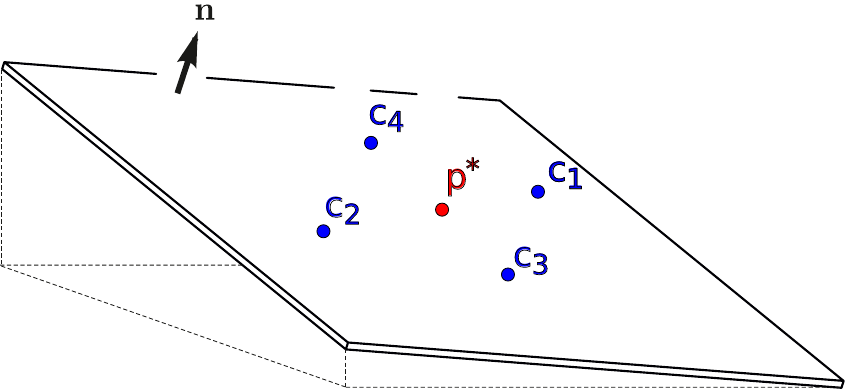}
\caption{Examples of making contact with the surface. 
      The red point is the desired position for the next footstep $\bp^*$,
      and the blue points are the potential contact points $\bc^1, \bc^2, \bc^3, \bc^4$ from exploration.
}
\label{f:exploration-contact-points} 
\end{figure}

\subsection{Estimation of surface normal}
\label{s:method-estimation}
\noindent 
After the exploration, we have a set of $\dimPoints$ points $\left[\bc_1, \bc_2, \dots, \bc_\dimPoints \right]$  on the contact surface $\contactSurface$.
We define $\contactVectorSet \in \R^{\dimFeet \times \dimVectors}$ be a set of  $\dimVectors ={\dimPoints \choose 2}$ vectors 
$\contactVectorSet=\left[\contactVectors_1, \contactVectors_2, ..., \contactVectors_v\right]$ such that
$\bv_{k} = \bc_i - \bc_j~\forall~i \neq j$ is a vector that connects two contact points (see~\fref{exploration-contact-vector}).

\begin{figure}
  \centering
  \includegraphics[width=.7\linewidth]{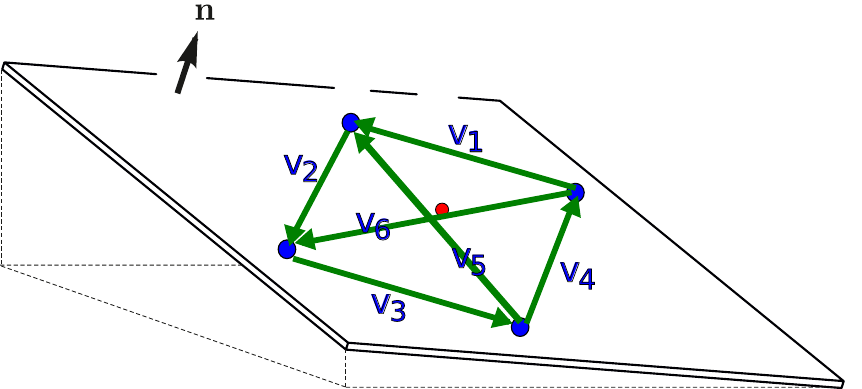}
  \caption{
  An example of contact vectors.
  The red point is the desired position for the next footstep $\bp^*$,
  and the blue points are the potential contact points $\bc^1, \bc^2, \bc^3, \bc^4$ from exploration.
  and the green vectors  $\contactVectorSet=\left[\contactVectors_1, \contactVectors_2, ..., \contactVectors_6\right]$  are the result of connecting the contact points. }
  \label{f:exploration-contact-vector} 
\end{figure} 

\begin{remark}
   $\contactVectorSet$ can be interpreted as a set of vectors that lie in the contact surface $\contactSurface$. 
  This also means, all vectors in $\contactVectorSet$ are orthogonal to the surface normal., i.e., $\bv_{k}^\T \contactNormal = 0,~ \forall \bv_k \in \contactVectorSet$.
  Therefore, we seek to solve the homogeneous linear equations 
\begin{equation}
  \contactVectorSet^\T \contactNormal  = \bZero.
 \label{e:estimation-objective}
\end{equation}
\end{remark}

\noindent Let $\contactVectorSet^\T= \bU \bS \bV^\T$ be the singular value decomposition of $\contactVectorSet$ such that $\bU$ is the matrix of left singular vectors, 
$\bS$ is a diagonal matrix of singular values, and
$\bV$ is a matrix of the right singular vectors.
Let $\bA \in \R^{3\times3}= \contactVectorSet \contactVectorSet^\T$ such that  
\begin{equation}
  \begin{aligned}
    \bA &=  \left(\bU \bS \bV^\T\right)^\T \left(\bU \bS \bV^\T\right)\\
	&= \bV ( \bS^\T \bS ) \bV^\T  
  \end{aligned} 
  \label{e:estimation-hermitaian}
\end{equation}

\begin{remark}
In~\eref{estimation-hermitaian}, we aim to make $\bA$ a Hermitaian matrix (i.e., $\bA \equiv \bA^\T$) such that 
the eigenvalues $\bS^\T \bS \in \R$ are always real, and it is always possible to find an orthogonal basis.
The surface normal $\contactNormal$ can be found by taking the columns of $\bV$ that corresponds to the smallest singular value.
\end{remark}

\noindent The rank of the constraint can be seen from the rank of $\bS^\T\bS$:
\begin{itemize}
 \item $rank(\bS^\T\bS)=1$: 
    The observations $\contactVectorSet$ spans a plane. 
    There is a 1-dimensional constraint and the robot can stand on a surface. 
    This observation means that $\bA$ has a vanishing singular value, and any linear combination of the corresponding right-singular vectors is a valid solution. 
    Since we are looking for a unit vector, the solution is the right-singular vector.
  
    \item $rank(\bS^\T\bS)=2$: There is a 2-dimensional constraint and the robot can stand on a line.
    If this is the case, we know the desired footstep position $\bp^*$ is not a good choice for the robot since the foot is likely to be stuck.
    This is a special case and out of the scope of this work. 
\end{itemize}

\subsection{Stopping Criterion}
\label{s:normal-probability}
\noindent 
In this section, we introduce a criterion to evaluate the confidence of the estimation.
Specifically, we would like to know how well our estimated contact normal $\contactNormal$ is without the knowledge of the true contact normal $\contactNormal^*$.
We will use the conjugate prior~\cite{Murphy.2012} to represent the unknown mean and variance of the ground truth, and then infer the probability of our estimation.

\subsubsection{Likelihood}
Given a set of observations $\contactVectorSet$; we would like to know what is the likelihood of this observation given the current estimate of contact normal. 
We assume that the inner product of $\bv$ and $\contactNormal$ has normal distribution $\bv^\T \contactNormal\sim \NormalDist{\obsAve}{\obsVar}$ where 
$\obsAve$ is the observation mean, and
$\obsVar$ is the observation variance.
The likelihood of the observations given the estimation $\prob{\contactVectorSet|\contactNormal}$ can be inferred from the probability density function 
\begin{equation}
\begin{aligned}
 \prob{\contactVectorSet|\contactNormal, \obsAve, \obsVar} &= \prod_k \frac{1}{\sqrt{2\pi\obsVar}} \exp^{-\frac{ (\bv_k^\T \contactNormal-\obsAve)^2}{2\obsVar}} \\
\end{aligned}
\end{equation}

\noindent 
If the estimation is correct, then~\eref{estimation-objective} is satisfied, so we can assume $\obsAve = 0$.
Let $\obsPre = \frac{1}{\obsVar}$ denotes the precision of the observations. The above equation can be re-written as
\begin{equation}
\begin{aligned}
 \prob{\contactVectorSet|\contactNormal,\obsPre} 
				    &= \frac{1}{(2\pi)^{n/2}} \obsPre^{n/2} \exp^{-\frac{\obsPre}{2} \sum_k ( \bv_k^\T \contactNormal )^2 }  
\end{aligned}
\label{e:normal-likelihood}
\end{equation}

\subsubsection{Prior distribution}
The conjugate prior of a normal distribution with unknown mean and variance can be represented by Normal-gamma distribution 
$\contactNormal \sim \mathcal{N}\mathcal{G}(\NormalGammaAve, \kappa, \alpha, \beta)$ where $\NormalGammaAve, \kappa, \alpha, \beta$ are the unknown hyper-parameters of the distribution.
The probability of $\contactNormal$ given $\obsPre$ is defined by
\begin{equation}
\begin{aligned}
\prob{\contactNormal,\obsPre| \NormalGammaAve_0, \kappa_0, \alpha_0,\beta_0} =&\\
  \frac{\beta_0^{\alpha_0}} {\varGamma(\alpha_0)}  
  \sqrt{\frac{\kappa_0}{2\pi} } 
  \obsPre^{\alpha_0-0.5} &
  \exp^{ - \frac{\obsPre}{2} \left[\kappa_0 \sum( \contactNormal-\NormalGammaAve_0)^2 + 2\beta_0 \right]}  
\end{aligned}
\label{e:normal-prior}
\end{equation}

\subsubsection{Initial values}

Initially, we assume the robot is going to walk over flat terrain, so the initial guess of the contact normal is set to $\NormalGammaAve_0=[0,0,1]^\T $ 
The parameters of the prior distribution are set to $\alpha_0 = 0.1, \beta_0= 0.1, \kappa_0=1$.

\subsubsection{Posterior distribution}
\noindent 
The likelihood of the estimation $\contactNormal_k$ given the observed data samples $\contactVectorSet$, or posterior distribution, 
can be written as $\prob{\contactNormal_k | \contactVectorSet} = \prob{\contactVectorSet|\contactNormal_k} \prob{\contactNormal_k}$. 
By the property of Normal-gamma distribution, 
if the prior is a Normal-gamma distribution $\contactNormal_k \NormalGammaDist{\NormalGammaAve_k, \kappa_k, \alpha_k, \beta_k}$, 
the posterior can be proven to be a Normal-gamma distribution  $\contactNormal_{k+1} \NormalGammaDist{\NormalGammaAve_{k+1}, \kappa_{k+1}, \alpha_{k+1}, \beta_{k+1}}$. 

\noindent 

\begin{remark}
By multiplying~\eref{normal-likelihood} and~\eref{normal-prior} and simplifying the expression, the posterior distribution can be described by another Normal-Gamma distribution. For each iteration, we only need to update the hyper-parameters of the distribution in order to evaluate the estimation.
\begin{equation}
  \begin{aligned}
  \NormalGammaAve_{k+1} &= \frac{\lambda_0 \mu_0 + k \contactNormal }{\kappa_0+k} \\
  \kappa_{k+1} &= \kappa_0 + k \\
  \alpha_{k+1} &= \alpha_{0} + \frac{k}{2} \\
  \beta_{k+1} &= \beta_0 + \frac{1}{2} \sum( \contactVectorSet\contactNormal)^2 + \left[ \frac{  \kappa_0  k\sum(\contactNormal-\NormalGammaAve_0)^2} { 2 \kappa_0+k} \right] 
  \end{aligned} 
\end{equation}
\end{remark}

\noindent 
As soon as the cumulative density is high enough, i.e., $ \prob{\contactNormal_k | \contactVectorSet}  \geq \stoppingThreshold$, we can stop the exploration.

\section{Surface Friction Estimation}	\label{s:friction}	\noindent After learning the surface normal, we can decompose the contact force into normal and tangential parts.
In this section, we describe a method to estimate the friction coefficient $\contactFriction$.

\subsection{Surface Friction Estimation}
\noindent 
Since we use the swing leg to perform the exploration and standing with the other three legs, the force exerted by the robot is the only force that acts on the contact surface.
Therefore, the action force (e.g., the force applied by the foot) is equal to the reaction force (e.g., the ground reaction force) but the opposite direction; specifically,
\begin{equation}
 -\expForce = \contactForce
 \label{e:friction-action-reaction}
\end{equation}
\noindent
where $\expForce \in \R^3$ is the force exerted by the robot at the swing foot.
We use $\expNormal$ and $\expTangent$ to denote the normal and tangential components of $\expForce$, and 
$\expForce = \expNormal + \expTangent$ (see~\fref{friction-explore}).
 
\begin{figure}
  \centering
  \includegraphics[width=0.3\textwidth]{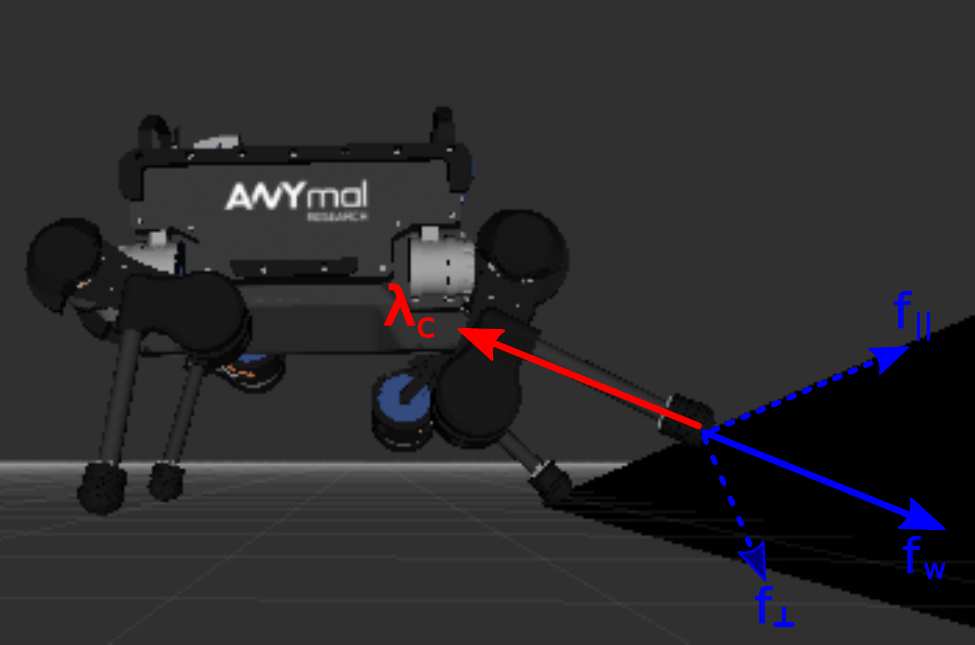}
  \caption{ 
  The force exerted by the robot $\expForce$ is divided into the normal component $\expNormal$ and the tangential component $\expTangent$.
  }
\vspace{-2mm}
  \label{f:friction-explore}
\end{figure}
 
During the last haptic exploration, we collected a set of contact points $\contactPoints_1, \contactPoints_2,..., \contactPoints_\dimPoints$. 
Since all these points are on the contact surface, we want to move the feet around these points. 
Namely, we generate the desired trajectory by interpolating the positions between all $\contactPoints_i$, and this trajectory will provide the desired foot position for the impedance controller 
\begin{equation}
  \bLambda \ddot{\tilde{\bp}} + \bD  \dot{\tilde{\bp}} + \bK \tilde{\bp}
\end{equation}
\noindent where
$\tilde{\bp} = \bp^*_w - \bp$ is the error between $\bc_i$ and the foot position, and 
$\bp^*_w$ denotes the desired foot position during exploration.
To ensure that the motion is moving on the contact surface, we project the motion onto the contact surface by
\begin{equation}
\expTangent = (\Identity_3 - \contactNormal^\T \contactNormal)  \left[  \bLambda \ddot{\tilde{\bp}} + \bD  \dot{\tilde{\bp}} + \bK \tilde{\bp} \right]
\end{equation}

\begin{remark}
Since $\contactNormal$ is a unit vector, the pseudo inverse is equal to its transpose. Therefore, the projection is defined by $\left( \Identity_3 - \contactNormal^\T \contactNormal \right)$ instead of $ \left( \Identity_3 - \contactNormal^\dagger \contactNormal \right)$
\end{remark}

\noindent From~\eref{background-friction}, we know that the foot cannot slide on the contact surface if $\mu \lambda_z \geq \sqrt{ \lambda_x^2 + \lambda_y^2 }$, and the vice versa.
Substituting~\eref{friction-action-reaction} into~\eref{background-friction} indicates that
\begin{equation} 
\begin{cases} 
   -\mu \expNormal \geq \vectornorm{ \expTangent } 	\rightarrow \text{~foot does not move} \\
   -\mu \expNormal < \vectornorm{ \expTangent }	 	\rightarrow  \text{~foot may move}  
\end{cases}
\label{e:friction-slippage}
\end{equation}

\noindent Therefore, we can vary the values of $\mu \in \left[0.2, 1.0\right]$ to get the normal force
\begin{equation}
\expNormal = \contactNormal \left( \frac{ \vectornorm{ \expTangent } } {\contactFriction } \right) 
\end{equation}
During the exploration, we start with a large friction coefficient. If there is a slippage, the friction coefficient will be reduced.
Finally, the total force at the swing foot will be $\expForce = \expNormal + \expTangent$, and the torque of the swing leg is $\expTorque = \bh_w + \expJacobian^\T \expForce$.

\subsection{Stopping Criteria}
\label{s:friction-probability}
\noindent 
During the exploration, we would like to know how likely our estimated friction coefficient $\mu$ is smaller or equal to the true friction value $\mu^*$, 
so we can stop exploration if we are confident about our estimation. 
For this, we will use the Bernoulli distribution with Beta distribution as the conjugate prior to~\cite{Bayesian.2012} model the probability distribution of our estimation.

\subsubsection{Likelihood distribution}
During the exploration, the swing foot is either sliding or non-sliding on the contact surface. 
We represent the outcome of the $k^{th}$ step as a boolean variable $\obsFriction^k \in \{0,1 \}$ where $\obsFriction^k=1$ denotes no-slippage and $\obsFriction^k = 0$ denotes slippage. 
Since we don't have the true friction coefficient, we cannot use~\eref{friction-slippage} to determine slippage; instead,
the slippage is determined by how much the desired end-effector velocity $\bv^* \neq \bZero$ is different from the observed end-effector velocity $\bv$ on the contact surface:

\begin{equation}
 \obsFriction = \begin{cases}
		  1, \text{ if } \vectornorm{ (\Identity_3 - \contactNormal^\T \contactNormal)  (\bv^*-\bv) } \geq \epsilon \\
		  0, \text{ if } \vectornorm{ (\Identity_3 - \contactNormal^\T \contactNormal)  (\bv^*-\bv) } < \epsilon 
              \end{cases} 
\label{e:friction-observation}
\end{equation}

\noindent Since the observation is binary, we represent the likelihood as Bernoulli distribution $\obsFriction^k | \contactFriction^k \sim Bernoulli(\prob{\contactFriction^k})$, and the probability density is defined as
\begin{equation}
    \prob{\obsFriction^k | \contactFriction^k} = \prob{\contactFriction^k}^{\obsFriction^k} (1-\prob{\contactFriction}) ^{1-\obsFriction^k}
\end{equation} 

\noindent Namely, if there is no slippage, the probability that the observation should not slip is the probability that the current friction coefficient setting is sufficient.

\subsubsection{Prior distribution}
\noindent 
The Beta distribution is the conjugate prior of the Bernoulli distribution. 
Although we do not have the true distribution of $\contactFriction$, the prior distribution can be described as 
\begin{equation}
    \contactFriction^k\sim \mathcal{B}(a^k,b^k) 
\end{equation}
where $\mathcal{B}$ denotes beta distribution and $a^k, b^k >0$ are  the unknown shape parameters of the distribution. The probability of the prediction $\prob{\contactFriction^k}$ can be estimated by the probability density function of beta distribution.

\subsubsection{Initial Guess}
The initial guess of the friction coefficient is $\mu^0 = 1.0$, since it is larger than friction coefficients in most scenarios.
These shape parameters are set to $a^0 = 1, b^0 =1$ at the beginning of the experiment. 
This is equivalent to a uniform distribution on the interval of 0 and 1 where all potential solutions have the same probability.

\subsubsection{Posterior distribution}
If the prior is of the form of a Beta distribution, then the posterior will be of the same distribution~\cite{Bayesian.2012}.
Therefore, the posterior distribution $\contactFriction^{k+1}$, given samples $\obsFriction^k$ also has a beta distribution 
\begin{equation}
    \contactFriction^{k+1} \sim  \mathcal{B}(a^k+\obsFriction^k, b^k+(1-\obsFriction^k)) 
\end{equation}

\begin{remark}
For each iteration, we only need to update the hyper-parameters of the distribution to evaluate the estimation.
\begin{equation}
  \begin{aligned}
   \alpha_{k+1} & = \alpha_k + \obsFriction \\
   \beta_{k+1} &= \alpha_k + (1-\obsFriction)
   \end{aligned} 
\end{equation}
\end{remark}

\noindent 
Then, we update the prior by substituting in the posterior and repeat. 
This process continues until we have enough confidence in the estimate. 
Specifically, we want to ensure the estimated friction coefficient is smaller than the true friction coefficient.
The estimation is completed if 
\begin{equation}
\prob{\contactFriction^* > \contactFriction^k} > \epsilon_\contactFriction 
\end{equation}
\noindent 
where  
$\prob{\contactFriction^* > \contactFriction^k}= 1 -\prob{\contactFriction^* < \contactFriction^k}\in \left[0,1\right]$, 
$\epsilon_\contactFriction$ is a predefined threshold function, and 
$\prob{\contactFriction^* \leq \contactFriction^k}$ is the cumulative distribution function of beta distribution.

\section{Evaluation}			\label{s:evaluation}	\noindent In this section, we describe how we validate our proposed method, in both simulation and on a real robotic platform.

\subsection{Learning surface normal}

\begin{figure}
  \centering
  \includegraphics[width=0.475\textwidth]{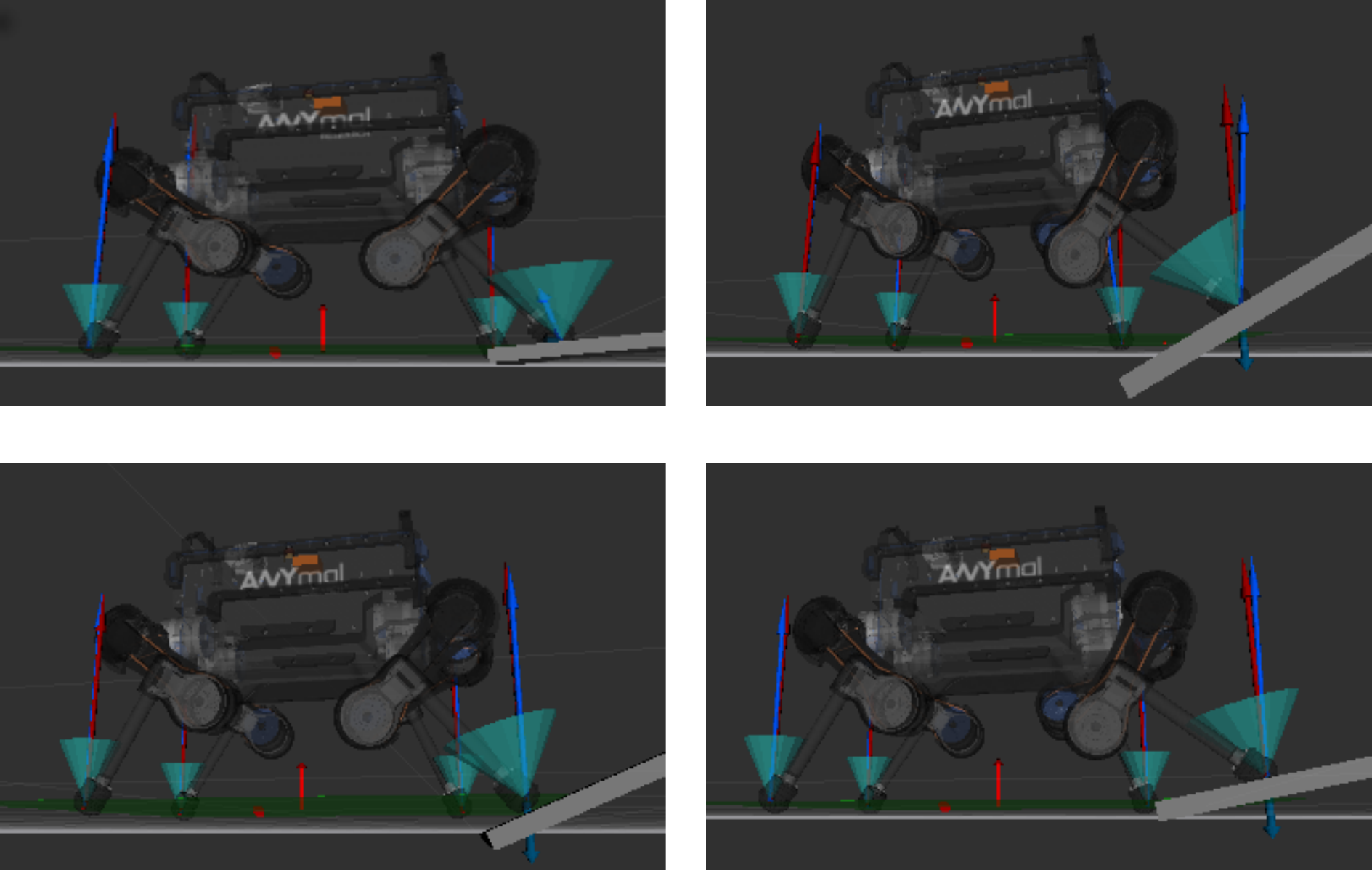}
  \caption{ 
  The result of estimating surface normal on different inclinations. We can see that the estimated cone is normal to the contact surface. 
  }
  \label{f:evaluation-normal-friction-cone}
\end{figure}
\noindent In the first experiment, we would like to demonstrate the idea in simulation.
First, we made different slopes in simulated scenarios, where the true inclination of the surface can be controlled.
We tested on the slope with inclination = $0.1, 0.2,...0.5$ radians.
The right front leg is used as a haptic sensor during the experiment.

On average, the mean and standard deviation of the prediction accuracy $\sum (\contactNormal - \contactNormal^*)^2 $ is $(5.61 \times 10^{-4}) \pm (6.63 \times10^{-4})$.
\fref{evaluation-normal-friction-cone} shows an example of the resulting friction cone using the estimated surface normal. 
We can see that the resulting friction cone is perpendicular to the contact surface, and the resulting contact force falls within this friction cone.

\subsection{Learning surface frictions}

\noindent In this experiment, we would like to test the accuracy of friction estimation in simulated scenarios. 
We generated terrains with different friction coefficients $\contactFriction^* \in [0.4, 0.8]$ and used the  method in~\sref{friction} to estimate the friction coefficient.

Overall, the mean and standard deviation of accuracy is $(3.08\pm 1.81)\times10^{-3}$ over 5 experiments. 
\fref{evaluation-friction-visualization} shows the resulting friction cones, from the left to right are $\contactFriction^*=0.4, 0.6, 0.8$. We can see that the shape of the friction cone changes as the friction coefficient gets bigger.

\fref{evaluation-friction-distribution} shows the progression of probability distribution over time. In this example, the true friction coefficient is $\contactFriction^* = 0.5$. 
In~\fref{evaluation-friction-distribution}\ref{f:evaluation-friction-pdf}, the x-axis is the friction coefficient, the y-axis is the probability density. The blue curve is the final estimation (at step $k$), and the red lines are the estimation at formal steps $k-10, k-20, k-30$. We can see that the posterior distribution of the estimate slowly converges to the unknown mean, and the variance decreases as we receive more observations. 

\begin{figure}
  \centering
  \includegraphics[width=0.475\textwidth]{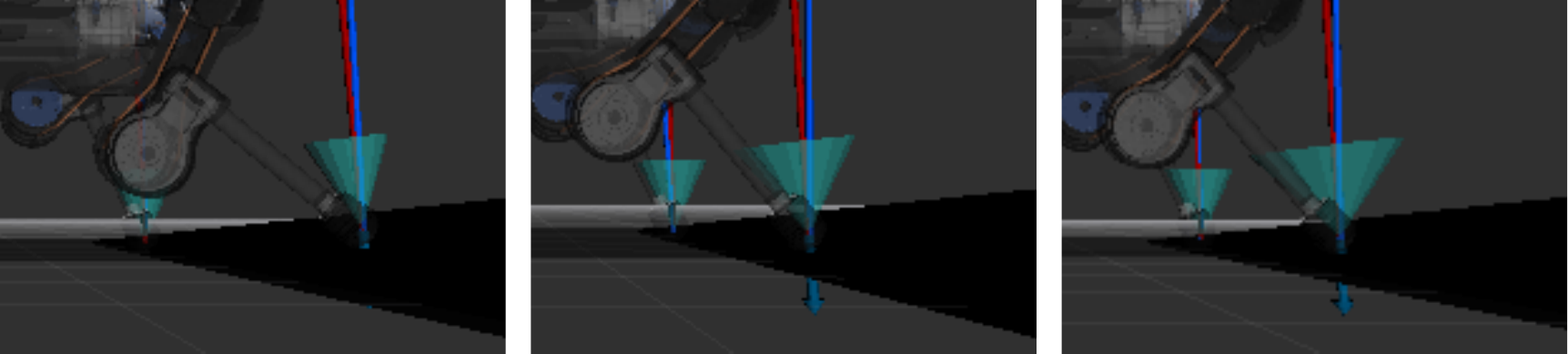}
  \caption{ 
  Resulting friction cones using the estimated friction coefficient. From the left to right are $\contactFriction^*=0.4, 0.6, 0.8$. 
  }
  \label{f:evaluation-friction-visualization}
\end{figure}

\fref{evaluation-friction-distribution}\ref{f:evaluation-friction-cdf} shows the cumulative distribution of the same example. 
In this figure, the x-axis is the iteration, the y-axis is the cumulative probability, and the red curve is the cumulative distribution over iterations. The blue line is the cut-off point where the cumulative distribution is good enough, indicates that the robot can stop exploration.

\begin{figure}[t!]
  \centering
   \begin{overpic} [width=.24\textwidth]{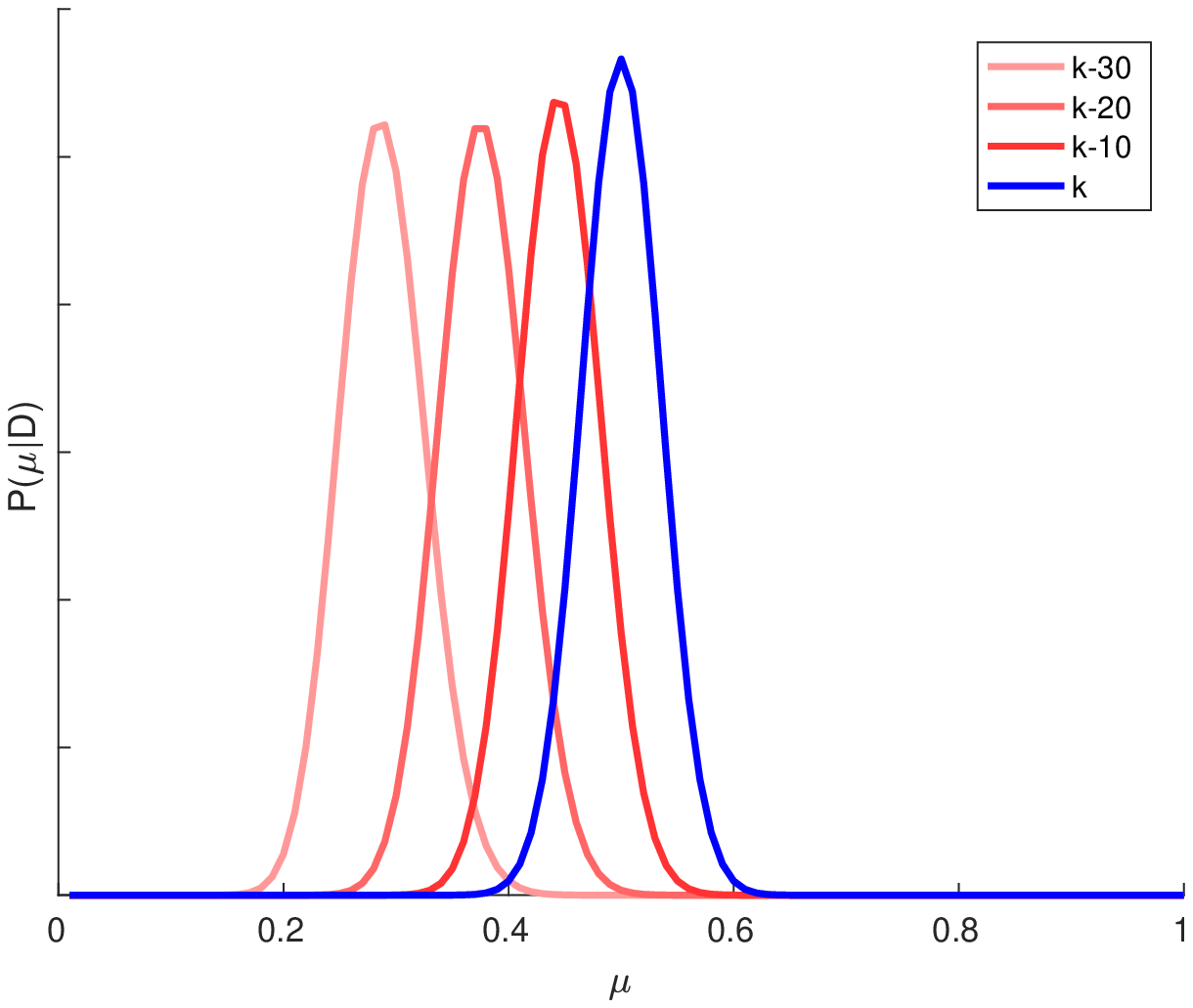}	{\manuallabel{f:evaluation-friction-pdf}{(a)}\color{black}\ref{f:evaluation-friction-pdf}} 	\end{overpic}~
   \begin{overpic} [width=.24\textwidth]{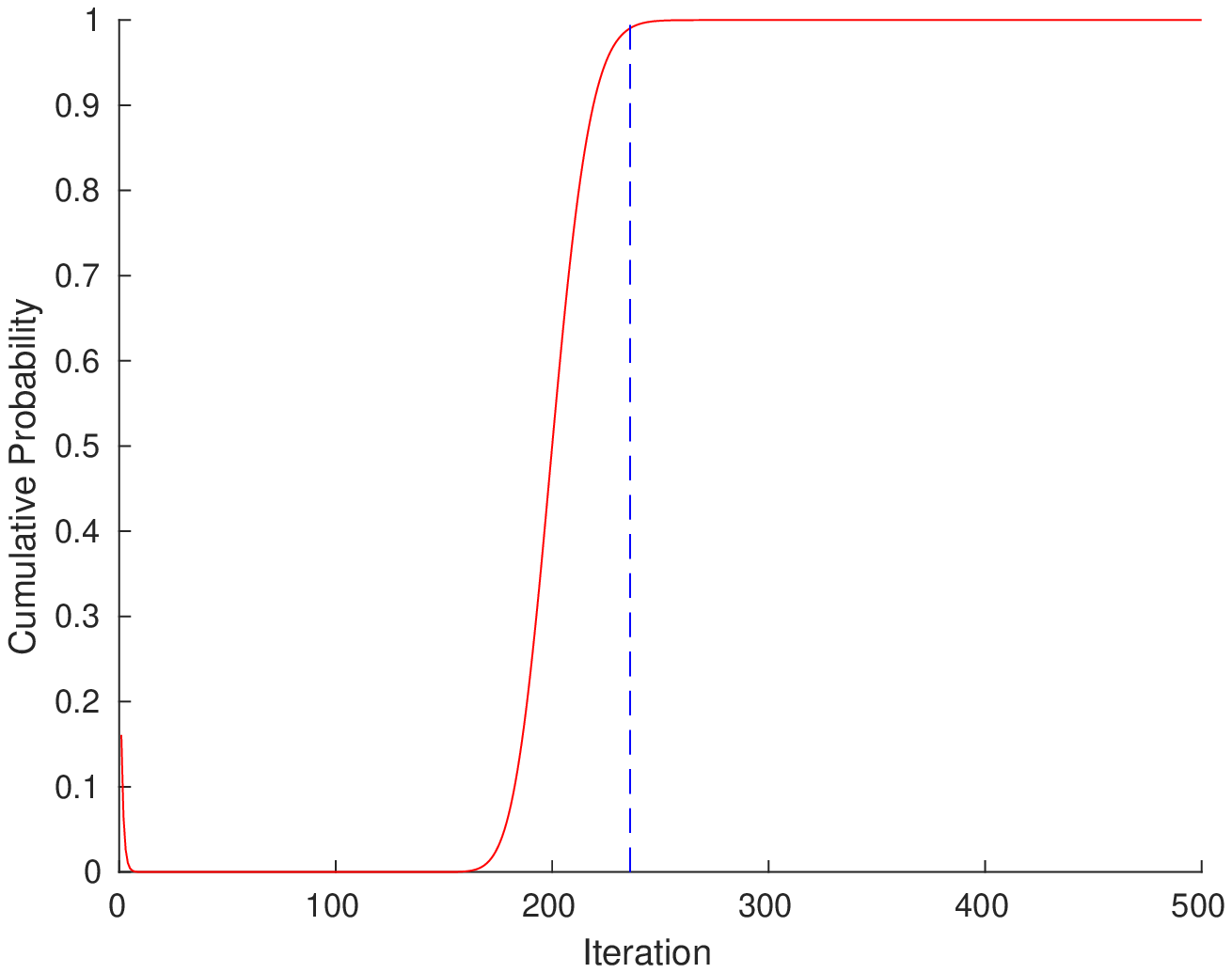}	{\manuallabel{f:evaluation-friction-cdf}{(b)}\color{black}\ref{f:evaluation-friction-cdf}} 	\end{overpic}~  
  \vspace{1mm} 
  \caption{ The progression of (a) posterior distribution and (b) cumulative distribution for estimating surface friction. The true friction coefficient is 0.5. }
  \label{f:evaluation-friction-distribution} 
\end{figure}

\subsection{Real Robotic Platform}
\noindent 
Finally, we conduct experiments using ANYmal~\cite{Hutter.2016}, a torque-controlled quadruped robot made by ANYbotics\footnote{See http://www.anybotics.com/}. 
The robot weights approximately 35 kg and has 12 joints actuated by Series Elastic Actuators (SEAs). Currently, the soft real-time control cycle is 2.5 ms. 
The control software is developed based on Robot Operating System (ROS). 

The terrain consisted of wedges with random inclination (see~\fref{evaluation-mockup-terrain}). Each wedge was about $50 \times 50 $ cm.
The robot used the proposed methods to estimate surface normal and friction before moving forward.

\begin{figure}
  \centering
  \includegraphics[width=0.27\linewidth, angle=-90]{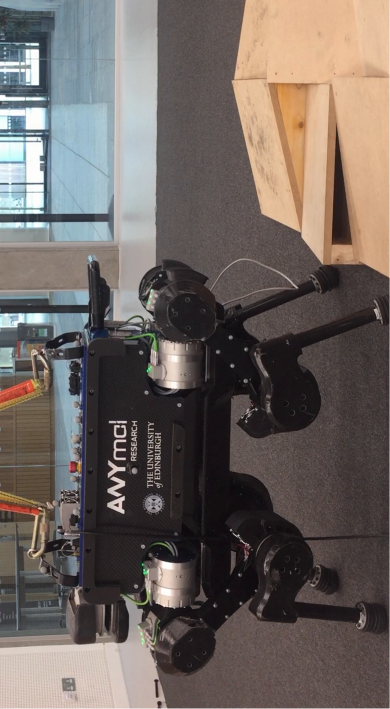}
  \includegraphics[width=0.27\linewidth, angle=-90]{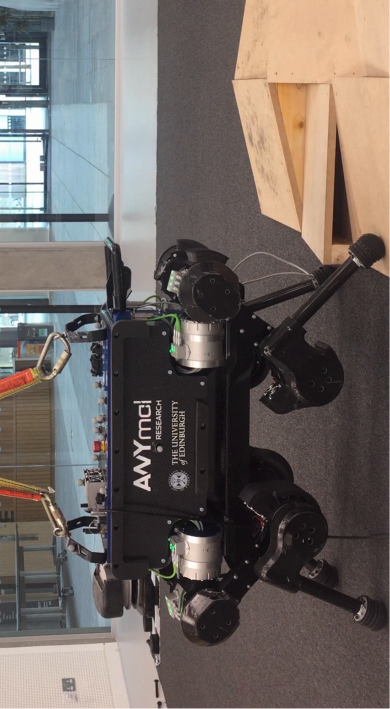}
  \caption{The mockup terrain with random inclinations}
  \label{f:evaluation-mockup-terrain}
\end{figure}

We do not have the ground truth of the surface normal, and the ground truth is approximated by measuring the dimensions of the wedges. 
The accuracy of predicting the surface normal is $8.02\times 10^{-3} \pm 1.01 \times 10^{-2}$. 
The friction coefficient is also not well defined, especially, it depends on the materials that interact. 
Therefore, we cannot evaluate the accuracy of the friction coefficient.
The evaluation is based on whether the robot can maintain contact with the terrain after the estimation.
Please see the supplementary video  \verb|https://youtu.be/SnafS_5361g| for the result on the real robotic platform.


\section{Conclusion}						\noindent 
This paper proposes a method for estimating surface information via haptic exploration.
We use one of the legs to act as a haptic sensor for exploring the environment before walking.
While collecting data, we estimate the surface normal and friction coefficient of the terrain, so the controller can optimize the contact force accordingly.
Probabilistic criteria are introduced to evaluate the quality of the estimate.
The method was validated on both a simulated environment and the quadruped robot ANYmal.

Future work will focus on incorporating vision data with the haptic exploration (e.g., explore the area where vision data are too noisy to evaluate) and applying this technique for footsteps planning. In addition, one issue we notice during the experiment is that the accuracy of contact detection (i.e., determining whether the robot is in contact or not) affects the accuracy of data collection. We will incorporate a better contact estimation~\cite{Mirrazavi.2018}\cite{Kim.2018.ICRA} with our haptic exploration.

\section*{Acknowledgement} 
\vspace{-1mm}
\noindent 
This work was funded by the European Commission Horizon 2020 Work Programme: THING ICT-2017-1 780883,

\bibliographystyle{bibtex/IEEEtran}
\bibliography{paper}


\end{document}